\documentclass[10pt,twocolumn,letterpaper]{article}

\usepackage[pagenumbers]{wacv} 

\usepackage{graphicx}
\usepackage{amsmath}
\usepackage{amssymb}
\usepackage{booktabs}

\usepackage[accsupp]{axessibility}
\usepackage{graphicx}
\usepackage{amsmath}
\usepackage{amssymb}
\usepackage{booktabs}
\usepackage[bb=boondox]{mathalfa}
\usepackage{nicefrac}
\usepackage{multirow}

\newcommand{\ddpm}[0]{\epsilon_\theta}
\newcommand{\E}[0]{\mathbb{E}}
\newcommand{\edit}[1]{#1\@\xspace}

\usepackage[pagebackref,breaklinks,colorlinks]{hyperref}

\usepackage[capitalize]{cleveref}
\crefname{section}{Sec.}{Secs.}
\Crefname{section}{Section}{Sections}
\Crefname{table}{Table}{Tables}
\crefname{table}{Tab.}{Tabs.}

\def\wacvPaperID{1129} 
\def\confName{WACV}
\def\confYear{2024}

\begin{document}

\title{Text-to-Image Models for Counterfactual Explanations: a Black-Box Approach}

\author{Guillaume Jeanneret,
Lo\"{i}c Simon,
Fr\'ed\'eric Jurie \\
Normandy University, ENSICAEN, UNICAEN, CNRS, GREYC, France \\
{\small \texttt{guillaume.jeanneret-sanmiguel@unicaen.fr}}}

\maketitle

\begin{abstract}
This paper addresses the challenge of generating Counterfactual Explanations (CEs), involving the identification and modification of the fewest necessary features to alter a classifier's prediction for a given image. Our proposed method, \textbf{T}ext-to-\textbf{I}mage \textbf{M}odels for Counterfactual \textbf{E}xplanations (TIME), is a black-box counterfactual technique based on distillation. Unlike previous methods, this approach requires solely the image and its prediction, omitting the need for the classifier's structure, parameters, or gradients. 
Before generating the counterfactuals, TIME introduces two distinct biases into Stable Diffusion in the form of textual embeddings: the context bias, associated with the image's structure, and the class bias, linked to class-specific features learned by the target classifier. 
After learning these biases, we find the optimal latent code applying the classifier's predicted class token and regenerate the image using the target embedding as conditioning, producing the counterfactual explanation. Extensive empirical studies validate that TIME can generate explanations of comparable effectiveness even when operating within a black-box setting.
\end{abstract}

\section{Introduction}
\label{sec:intro}

Recently, deep neural networks (DNN) have seen increased attention for their impressive forecasting abilities. 
The use of deep learning in critical applications, such as driving automation, made the scientific community increasingly involved in what a model is learning and how it makes its predictions. 
These concerns shed light on the field of Explainable Artificial Intelligence (XAI) in an attempt to ``open the black-box'' and decipher its induced biases. 


Counterfactual explanations (CEs) are an attempt to find an answer to this previous problem. 
They try answering the following question: \emph{What do we need to change in $X$ to change the prediction from $Y$ to $Z$}?
Because CEs give intuitive feedback about what to change to get the desired result, two applications use these explanations: feedback recommendation systems and debugging tools. 
Take an automated loan approval system as an example. 
From a user's point of view, if it gets a negative prediction, the user would be more interested in knowing what plausible changes can be made to get a positive result, rather than having an exhaustive list of explanations for why the result is unfavorable. 
From the debugger's point of view, it can look for biases that were considered in the decision when they should not have been, thus revealing the classifier's weaknesses.

\begin{table}[t]
    \centering
    \footnotesize
    \begin{tabular}{c|cccc} \toprule
       Method & Model & Training & Specificity & Optim. \\ \midrule
       DiVE~\cite{rodriguez2021beyond}   & VAE   & Days     & Only DNN    & Yes \\
       STEEX~\cite{jacob2022steex}  & GAN   & Days     & Only DNN    & Yes \\
       DiME~\cite{jeanneret2022diffusion}   & DDPM  & Days     & Only DNN    & Yes \\
       ACE~\cite{jeanneret2023adversarial}    & DDPM  & Days     & Only DNN    & Yes \\ \midrule
       TIME (Ours) & T2I & Hours     & Black-Box & No \\ \bottomrule
    \end{tabular}
    \caption{\textbf{Advantages of the proposed methodology.} TIME uses a pre-trained T2I model and trains only a few textual embeddings, requiring hours of training instead of days. It does not require access to the target model (completely black-box) and does not involve any optimization during counterfactual generation.}
    \label{tab:advantages}
    \vspace{-5mm}
\end{table}

While there are multiple ways to address this question for visual systems, \eg by adding adversarial noise~\cite{goodfellow2014explaining}, the modifications must be sparse and comprehensive to provide insight into which variables the model is using. 
To this end, most studies for CEs use generative models, such as GANs~\cite{NIPS2014_5ca3e9b1}, Denoising Diffusion Probabilistic Models (DDPMs)~\cite{ho2020denoising}, or VAEs~\cite{kingma2013auto}, as they provide an intuitive interface to approximate the image manifold and constrain the generation in an appropriate space. 
Although they have several advantages, training these generative models is cumbersome and may not yield adequate results, especially when the data is limited~\cite{karras2020training}. 
To this end, we expect that the use of large generative models trained on colossal datasets, such as LAION-5B~\cite{schuhmann2022laionb}, can provide a sufficient tool to generate CEs. 
On the one hand, these generative models have shown remarkable qualitative performance, an attractive feature to exploit. 
Second, since the generative model is already optimized, it can be used to capture data set specific concepts - \eg textual inversion~\cite{gal2023an} captures the main aspects of a target object when subject to only three to five images. 

In this paper, we explore how to take advantage of Text-to-Image (T2I) generative models for CEs - specifically, using Stable Diffusion~\cite{esser2020taming}. 
To do so, we take a distillation approach to transfer the learned information from the model into new text embeddings to align the concept class in text space. 
Second, we use inversion techniques~\cite{wallace2023edict} to find the optimal noise to recover the original instance.
Finally, with our distilled knowledge, we denoise this optimal point to recover the final instance using the target label, thus generating the CE.
This is advantageous because we can tackle the challenging scenario of explaining a black-box model, \ie having access only to its predictions.

Our proposed approach has three main advantages over previous literature, as shown in Table~\ref{tab:advantages}. 
First, we only train some textual embeddings, making the training efficient, while previous methods require training a generative model from scratch.
Second, we do not require an optimization loop when generating the final counterfactual, which reduces the generation time. 
Finally, our explainability tool works in a completely black-box environment. 
While most modern approaches~\cite{rodriguez2021beyond,jacob2022steex,jeanneret2022diffusion,zemni2023octet,jeanneret2023adversarial} are DNN-specific, because they rely on gradients, our approach, which uses only the output and input as cues, can be used to diagnose any model regardless of its internal functioning. 
This setting is crucial for privacy-preserving applications, such as medical data analysis, since eliminating access to the gradients could prevent data leakage~\cite{NEURIPS2019_60a6c400}, as it helps protect personal or confidential information.

We summarize our contributions as follows\footnote{\edit{Code is available at \href{https://github.com/guillaumejs2403/TIME}{https://github.com/guillaumejs2403/TIME}}}:
\begin{itemize}
    \item We propose TIME: Text-to-Image Models for Counterfactual Explanations, using Stable Diffusion~\cite{Rombach_2022_CVPR} T2I generative model to generate CEs.
    \item Our proposed approach is completely black-box. 
    \item Our counterfactual explanation method based on a distillation approach does not require any optimization during inference, unlike most methods.
    \item From a quantitative perspective, we achieve similar performance to the previous state-of-the-art, while having access only to the input and the prediction of the target classifier.
\end{itemize}

\section{Related Work}

\subsection{Explainable Artificial Intelligence}
The research branch of XAI broads multiple ways to provide insights into what a model is learning. 
As a bird's view analysis, there are two main distinctions between methods: \emph{Interpretable by-design} architectures, and \emph{Post-Hoc} explainability methods. 
The former searches to create algorithms that directly expose why a decision was made~\cite{nauta2023pipnet,donnelly2022deformable,NEURIPS2019_adf7ee2d,Bohle_2022_CVPR,Bohle_2021_CVPR,Huang_2020_CVPR,zhang2018interpretable}. 
Our research study is based on the latter. 
\textit{Post-hoc} explainability methods study pretrained models and try to decipher the variables used for forecasting. 
Along these lines, there are saliency maps~\cite{Jung_2021_ICCV,Petsiuk2018rise,8354201,8237336}, concept attribution~\cite{kim2018interpretability,fel2023craft,ghorbani2019towards}, or distillation approaches into interpretable by-design models~\cite{ge2021peek}. 
In this paper, we study the on-growing branch of CEs~\cite{wachter2018CounterfactualExplanationsOpening}. 
In contrast to previous methods, these explanations are simpler and more aligned with human understanding, making them appealing to comprehend machine learning models.

\subsection{Counterfactual Explanations}

The seminal work of Watcher~\etal~\cite{wachter2018CounterfactualExplanationsOpening} defined what a counterfactual explanation is and proposed to find them as a minimization problem between a classification loss and a distance loss. 
In the image domain, optimizing the image's raw pixels produces adversarial noises~\cite{goodfellow2014explaining}. 
So, many studies based their work on Watcher~\etal~\cite{wachter2018CounterfactualExplanationsOpening}'s optimization procedure with a generative model to regularize the CE production, such as variational autoencoders~\cite{rodriguez2021beyond}, generative adversarial networks~\cite{jacob2022steex,zemni2023octet,khorram2022cycle,luo2023zero,Singla2020Explanation}, and diffusion models~\cite{jeanneret2022diffusion,Sanchez2022DiffusionCM,jeanneret2023adversarial,augustin2022diffusion}. 
In contrast to these works, our proposed approach, TIME, is a distillation approach for counterfactuals. 
Our method does not require any optimization loop when building the explanation, since we transfer the learning into the T2I model. 
Furthermore, we do not require access to the gradients of the target model but only the input and output, making it black-box, unlike previous methods.

Co-occurent works analyze dataset biases using T2I models to create distributional shifts in data~\cite{Vendrow2023DatasetID,prabhu2023bridging}. 
Although a valid approach to debug datasets, we argue that these approaches do not search what a model learned but instead a general strategy for the biases in datasets under distributional shifts (\eg it is normal to misclassify a dog with glasses since the model was not trained to classify dog with glasses). 
Further, their proposed approaches are computationally heavy, since they require fine-tuning Large Language Models or optimizing each inversion step on top of Stable Diffusion. 
Instead, ours requires training a word embedding, and the inference merely requires Stable Diffusion without computing any gradients, which fits into a single small GPU.

\subsection{Customization with Text-to-Image Models}

Due to the interest in creating unimaginable scenarios with personalized objects, customizing T2I diffusion models has gained attention in recent literature. 
Textual Inversion~\cite{gal2023an} and following works~\cite{Ruiz_2023_CVPR,dong2023prompt,gu2023photoswap,zhang2023adding,mou2023t2i} are popular approaches to learn to generate specific objects or styles by fine-tuning all or some part of the T2I model. 
Thus, the new concept can be used in a phrase such that the T2I model will synthesize it. 

One of the most difficult problems is editing real-world images with T2I models. The pioneer work of Song \etal~\cite{song2020denoising} proposed a non-stochastic variant of DDPMs, called Denoising Diffusion Implicit Models (DDIM). Hence, a single noise seed yields the same image. 
So, to find an approximate noise, DDIM Inversion noises the image using the diffusion model. 
Yet, some problems arise with this approximation. 
So, novel works~\cite{mokady2023null,parmar2023zero} modify the inversion process by including an inner gradient-based optimization at each noising step, making it unfeasible when analyzing a bundle of images. 
Finally, Wallace~\etal~\cite{wallace2023edict} proposed to modify the DDIM algorithm into a two-stream diffusion process, reaching a ``perfect'' inversion. 
We take advantage of these works and distill the learned information from a classifier to generate counterfactual explanations of real images, a step to interpret the target classifier.
\section{Methodology}

This section explains the proposed methodology for generating counterfactuals using T2I generative models. 
In section~\ref{sec:prel}, we briefly introduce some useful preliminary concepts of diffusion models. 
Then we describe our proposed method in a three-step procedure. 
First, we explain how to transfer what the classifier has learned into the generative model as a set of new text tokens (Section~\ref{sec:distill}). 
Second, using recent advances in DDIM Inversion, we revert the image to its noise representation using the original prediction of the classifier. 
Finally, we denoise the noisy latent instance using the target label (Section~\ref{sec:gen}).

\subsection{DDPM Preliminaries}\label{sec:prel}

Diffusion models~\cite{ho2020denoising} are generative architectures that create images by iteratively \emph{removing} noise. 
DDPMs are based on two inverse Markov chains. 
The forward chain \emph{adds} noise, while the reverse chain \emph{removes} it. 
Thus, the generation process is reverse denoising, starting from a random Gaussian variable and removing small amounts of noise until a plausible image is returned.

Formally, given a diffusion model $\ddpm$ and a fixed set of steps $T$, $\ddpm$ takes as input a noisy image $x_t$, the current step $t$ to compute a residual shift, and a textual conditioning $C$, in our case. 
For the generation, $\ddpm$ updates $x_t$ following:
\begin{equation}\label{eq:sampling}
    x_{t-1} = \frac{1}{\sqrt{\alpha_t}} \left(x_t - \frac{1 - \alpha_t}{\sqrt{1 - \Bar{\alpha}_t}}\ddpm(x_t, t, C)\right) + \sigma_t \epsilon,
\end{equation}
where $\sigma_t$, $\alpha_t$ and $\Bar{\alpha}_t$ are some predefined constants, and $\epsilon$ and $x_T$ are extracted from a Gaussian distribution. 
This process is repeated until $t=0$. 
To train a DDPM, for a given an image-text pair $(x, C)$, each optimization step minimizes the loss:
\begin{equation}\label{eq:training}
    L(x, \epsilon, t, C) = \| \epsilon - \ddpm(x_t(x, t, \epsilon), \, t, C)\|^2,
\end{equation}
with
\begin{equation}
    x_t(x, t, \epsilon) = \sqrt{\Bar{\alpha}_t}\,x + \sqrt{1 - \Bar{\alpha}_t}\,\epsilon.
\end{equation}

The pioneering work of Ho~\etal\cite{ho2020denoising} focused on training and evaluating these models in the pixel space, making them computationally heavy. 
Latent Diffusion Models~\cite{Rombach_2022_CVPR} proposed to reduce this burden by performing the diffusion process in the latent space of a Quantized Autoencoder~\cite{esser2020taming}.
Further, they augment the generation by using textual conditioning $C$ at its core to steer the diffusion process, as well as increasing the quality of the generation using Classifier-Free Guidance~\cite{ho2021classifierfree} (CFG). 

The CFG~\cite{ho2021classifierfree}'s core modifies the sampling strategy in Eq.~\ref{eq:sampling} by replacing $\ddpm$ with $\ddpm^f$, a shifted version defined as follows: 
\begin{equation}\label{eq:class-free}
\begin{split}
    \ddpm^f(x_t, t, C) := (1 + w)\, \ddpm(x_t, t, C) - w\, \ddpm(x_t, t, \varnothing),
\end{split}
\end{equation}
where $\varnothing$ is the empty conditioning and $w$ is a weighting constant, resulting in a qualitative improvement. 


\subsection{Distilling Knowledge into Stable Diffusion}\label{sec:distill}

To use large generative models, and in particular Stable Diffusion~\cite{Rombach_2022_CVPR}, we chose to distill the learned biases of the target classifier into the generative model to avoid \edit{any gradient-based optimization during the CE formation.} 

A model is subject to several biases as it learns, of which we distinguish two. 
The first is a \emph{context bias}. 
This bias refers to the way images are formed. 
For example, ImageNet images~\cite{deng2009imagenet} tend to have the object (\eg, animals, cars, bridges) in the center, while CelebA HQ images~\cite{CelebAMask-HQ} are human faces.
The second bias is class-specific, and it relates to the semantic cues extracted by the classifier to make its decision, \eg white and black stripes for a zebra.

So, we take a textual inversion approach to distill the context bias and the knowledge of the target classifier into the textual embedding space of Stable Diffusion. 
In a nutshell, textual inversion~\cite{gal2023an} links a new text-code $c^*$ and an object (or style) such that when this new code is used, the generative model will generate this new concept. 
To achieve this, Gal~\etal~\cite{gal2023an} proposed to instantiate a new text embedding $e^*$, associate it to the new text-code $c^*$, and then train $e^*$ by minimizing the loss
\begin{equation}\label{eq:min}
    \E_{(x, C)\sim D, t\sim U[1,T], \epsilon\sim\mathcal{N}(0,I)}\left[L(x, t, \epsilon, C)\right].
\end{equation}
Here, $D$ is the set of images containing the concept to be learned, $U$ is the uniform distribution of natural numbers between $1$ and $T$, and $C$ is a text prompt containing the new text code $c^*$. 

Accordingly, to distill the context bias into Stable Diffusion, we follow~\cite{gal2023an} practices and learn a new textual embedding $e^*_{context}$ minimizing Eq.~\ref{eq:min} using as the conditioning the phrase \texttt{A $c^*_{context}$ picture}. 
Here, $c^*_{context}$ is the textual code related to textual embedding $e^*_{context}$.
In our setup, we used the complete training set of images with no labels where the model was trained. 

So far, we have not been required to use the classifier. 
To transfer the knowledge learned by the classifier to the T2I generation pipeline, we follow a similar approach. 
In this case, we train a new textual embedding $e^*_{i}$ for each class $i$ and represent its text token with $c^*_{i}$. 
However, instead of using the full training dataset $D$, we used only those images that the classifier predicted to be the source class $i$. 
As for the conditioning sentence, we take the previously learned context token and add the new class token to the sentence. 
Thus, we optimize Eq.~\ref{eq:min} with the new phrase \texttt{A $c^*_{context}$ image with a $c^*_{i}$} and the filtered dataset. 
\edit{For the rest of the text, we will refer to this prompt as $C_i$.}

\subsection{Counterfactual Explanations Generation}\label{sec:gen}

Now we want to use the learned embeddings to generate explanations. 
Current research on diffusion models has attempted to recover input images by retrieving the best noise, such that when the DDIM sampling strategy is used, it generates the initial instance. 
This is advantageous for our goal, since we can use current technological advances to generate this optimal latent noise and then inpaint the changes necessary to flip the classifier.

Since we need to perform perfect recovery to avoid most changes in the input image, we use EDICT~\cite{wallace2023edict}'s perfect inversion technique. 
In fact, they showed that inverting an image with a caption (Eqs~\ref{eq:invert}) and then denoising it (Eqs.~\ref{eq:denoise}) with a modified version of the original caption will produce semantic changes in the image.
In short, EDICT modifies the DDIM~\cite{song2020denoising} sampling strategy for diffusion models into a two-flow invertible sequence. 
By introducing a new hyperparameter $0<p<1$, setting $x_0$ and $y_0$ as the target image, and new variables:
\begin{equation}
    \begin{split}
        a_t &= \sqrt{\Bar{\alpha}_{t-1} / \Bar{\alpha}_t} \\
        b_t &= \sqrt{1 - \Bar{\alpha}_{t -1}} - \sqrt{\Bar{\alpha}_{t-1}(1-\Bar{\alpha}_t) / \Bar{\alpha}_t},
    \end{split}
\end{equation}
the denoising phase becomes:
\begin{equation}\label{eq:denoise}
    \begin{split}
        x_t^{inter} &= a_t\, x_t + b_t\, \ddpm^f(y_t, t, C)\\
        y_t^{inter} &= a_t\, y_t + b_t\, \ddpm^f(x_t^{inter}, t, C)\\
        x_{t-1} &= p\, x_t^{inter} + (1 - p)\, y_t^{inter} \\
        y_{t-1} &= p\, y_t^{inter} + (1 - p)\, x_{t - 1}.
    \end{split}
\end{equation}
In a similar vein, the inversion phase is the inverse of Eqs.~\ref{eq:denoise}:
\begin{equation}\label{eq:invert}
    \begin{split}
        y_{t+1}^{inter} &= (y_t - (1 - p)\, x_t)\; /\; p\\
        x_{t+1}^{inter} &= (x_t - (1 - p) \,y_{t+1}^{inter}) \; / \; p\\
        y_{t+1} &= \frac{1}{a_{t+1}} (y_{t+1}^{inter} - b_{t+1}\, \ddpm^f(x_{t+1}^{inter}, t+1, C))\\
        x_{t+1} &= \frac{1}{a_{t+1}} (x_{t+1}^{inter} - b_{t+1}\, \ddpm^f(y_{t+1}^{inter}, t+1, C)).
    \end{split}
\end{equation}

We can see a clear connection between Wallace~\etal~\cite{wallace2023edict}'s work and our main objective. 
If we invert an image using the caption with our context and source class tokens and then denoise it by changing the prompt to include the target token (learned in Section~\ref{sec:distill}), we can hope to generate the necessary changes to flip the classifier's decision. 

However, while adapting the EDICT method, we noticed a major problem with this approach. 
Although the chosen algorithm recovers the input instance, many images were difficult to modify. 
To circumvent this issue, we had to adjust the scores of the CFG in Eq.~\ref{eq:class-free}. 
As diffusion models are seen as score-matching models, the term 
\begin{equation}
    w(\ddpm(x_t, t, C_i) - \ddpm(x_t, t, \varnothing))
\end{equation}
in Eq.~\ref{eq:class-free} are gradients pointing to the target distribution conditioned on $C_i$. 
\edit{We call this the positive drift.}
Thus, by including a negative drift term,
\begin{equation}\label{eq:neg-grads}
    -w(\ddpm(x_t, t, C_j) - \ddpm(x_t, t, \varnothing)),
\end{equation}
we can lead the generation process \emph{away} from the source distribution conditioned in $C_j$. 
Therefore, we reformulate the CFG scores $\ddpm^f$, and rename it to $\ddpm^c$, as follows:
\begin{equation}\label{eq:drift}
\begin{split}
    \ddpm^c(x_t, t, C_i, C_j) = (1&+w)\,\ddpm(x_t, t, C_i) \\
    &- w\,\ddpm(x_t, t, C_j).
\end{split}
\end{equation}

\begin{figure}[t]
    \centering
    \includegraphics[width=0.90\columnwidth]{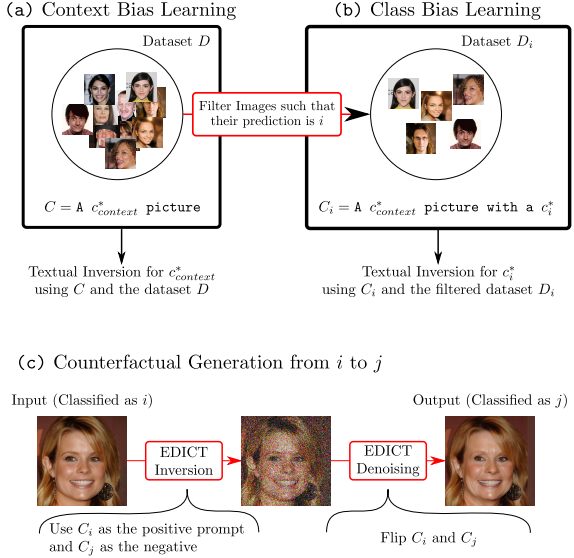}
    \caption{\textbf{TIME Overview.} Our proposed method consists of three steps: (a) We learn a context token for the whole dataset using textual inversion. (b) We filter out the images that the classifier predicts as source class $i$ and learn a new embedding. (c) Finally, to generate the counterfactual explanation, we invert the input image using a prompt containing the source embedding and then denoise it using the target embedding.}
    \label{fig:main-figure}
    \vspace{-3mm}
\end{figure}

As a result, and given the previously introduced notions, we propose \textbf{T}ext-to-\textbf{I}mage \textbf{M}odels for counterfactual \textbf{E}xplanations (TIME), illustrated in Figure~\ref{fig:main-figure}. 
To leverage these big generative models, we first distill the context bias into the pipeline's text embedding space by training a text embedding with the complete dataset. 
Then, we transfer the knowledge of the classifier by training a new embedding but using solely the instances with the same predictions. 
\edit{Finally, given an input image classified as $i$ and the target $j$, we invert the image (Eqs.~\ref{eq:invert}) using $\ddpm^c$ as the score network (Eq.~\ref{eq:drift}) using as the positive and negative drift $C_i$ and $C_j$, respectively.}
Then, we denoise the noisy state using Eqs.~\ref{eq:denoise} but switching textual conditionings.

\paragraph{Practical considerations.}
To avoid large changes in the image, the inversion stops at an intermediate step $\tau$ instead of $T$.
In addition, we have found that using more than a single embedding for the context and class biases yield further expressiveness. 
Also, if we fail to find a valid counterfactual, we choose a new $\tau$ and $w$ to rerun the algorithm. 
We will give the implementation details later in Section~\ref{sec:exp}.

\section{Experimental Validation}~\label{sec:exp}

\paragraph{Datasets and Models.} 
We evaluate our counterfactual method in the popular dataset CelebAHQ~\cite{CelebAMask-HQ}. 
The task at hand is classifying smile and age attributes from face instances, computed with a DenseNet121~\cite{huang2017densely} with an image resolution of $256\times256$ as in~\cite{jacob2022steex,jeanneret2023adversarial}. 
The evaluation is performed on the test set.
To make the assessment fair with previous methods, we used the publicly available classifiers for CelebA HQ dataset from previous studies~\cite{jacob2022steex}. 

\paragraph{Implementation Details.}
We based our approach on Stable Diffusion V1.4~\cite{esser2020taming}. For all dataset, we trained three textual embeddings for the context and class biases for 800 iterations with a learning rate of 0.01, a weight decay of 1e-4, and a batch size of 64. 
For the inference, we used the default EDICT's hyperparameter $p=0.93$ and a total of 50 steps. 
For the smiling attribute, we begin the CE generation with $(\tau, w) = (25, 3)$. In case of failure, we increased the tuple to $(30, 4), (35, 4)$ or $(35, 6)$. 
For the age attribute, we used $(\tau, w) \in \{(30, 4), (30, 6), (35, 4), (35, 6)\}$.
We performed all training and inference in a Nvidia GTX 1080.

\begin{figure*}[t]
    \centering
    \includegraphics[width=0.93\textwidth]{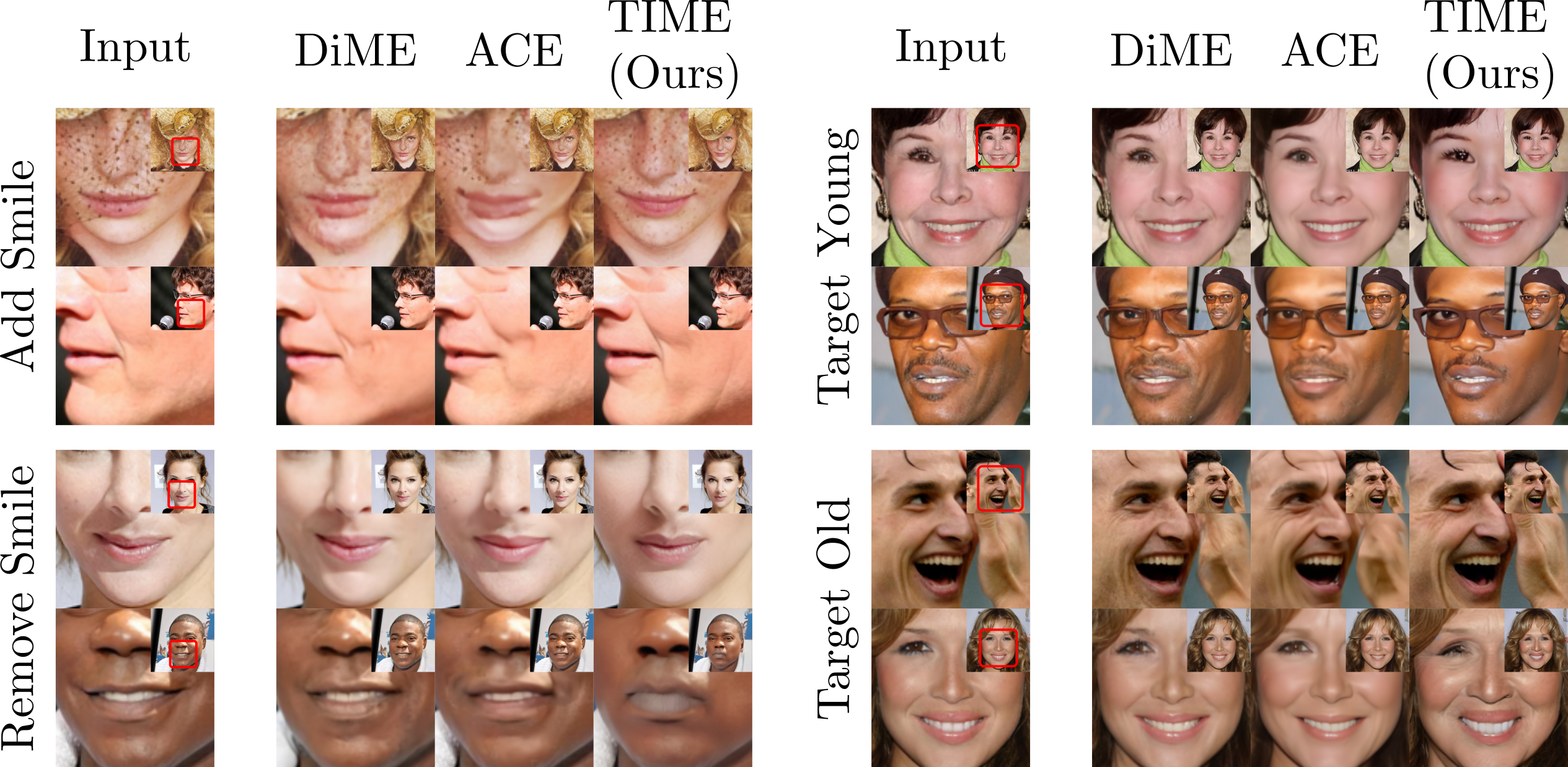}
    \caption{\textbf{Qualitative Results.} We present qualitative examples and compare them to the previous state of the art. DiME generates some out-of-distribution noise, while ACE creates blurry image sections. In contrast, TIME produces more realistic changes by harnessing the generative power of the T2I model.}
    \label{fig:qualitative}
    \vspace{-1mm}
\end{figure*}

\begin{table*}[t]
    \centering
    \resizebox{0.95\textwidth}{!}{
    \begin{tabular}{c|cccccccc|cccccccc} \toprule
        \multirow{2}{*}{Method} & \multicolumn{8}{|c|}{\textbf{Smile}} & \multicolumn{8}{c}{\textbf{Age}}\\\cmidrule{2-15}
                                                      &  FID ($\downarrow$)  & sFID ($\downarrow$) & FVA ($\uparrow$)  & FS ($\uparrow$)        & MNAC ($\downarrow$)  & CD ($\downarrow$)    & COUT ($\uparrow$) & SR ($\uparrow$) &  FID ($\downarrow$)  & sFID ($\downarrow$) & FVA ($\uparrow$)  & FS ($\uparrow$)        & MNAC ($\downarrow$)  & CD ($\downarrow$)    & COUT ($\uparrow$) & SR ($\uparrow$) \\ \midrule
        DiVE~\cite{rodriguez2021beyond}               & 107.0 & -    & 35.7 & -         & 7.41  & -     & -      & -    &107.5 & -    & 32.3 & -         & 6.76 & -    & -  & - \\ 
        STEEX~\cite{jacob2022steex}                   &  21.9 & -    & 97.6 & -         & 5.27  & -     & -      & -    & 26.8 & -    & 96.0 & -         & 5.63 & -    & -  & - \\
        DiME~\cite{jeanneret2022diffusion}            &  18.1 & 27.7 & 96.7 & 0.6729    & 2.63  & 1.82  & 0.6495 & 97.0 & 18.7 & 27.8 & 95.0 & 0.6597    & 2.10 & 4.29 & 0.5615 & 97.0 \\
        ACE* $\ell_1$~\cite{jeanneret2023adversarial} & 26.1  & 36.8 & 99.9 & 0.8020    & 2.33  & 2.49  & 0.4716 & 95.7 & 24.6 & 38.0 & 99.6 & 0.7680    & 1.95 & 4.61 & 0.4550 & 98.7 \\
        ACE $\ell_1$~\cite{jeanneret2023adversarial}  &  3.21 & 20.2 & 100.0& 0.8941    & 1.56  & 2.61  & 0.5496 & 95.0 & 5.31 & 21.7 & 99.6 & 0.8085    & 1.53 & 5.4  & 0.3984 & 95.0 \\
        ACE* $\ell_2$~\cite{jeanneret2023adversarial} & 26.0  & 35.2 & 99.9 & 0.8010    & 2.39  & 2.40  & 0.5048 & 97.9 & 24.2 & 34.9 & 99.4 & 0.7690    & 2.02 & 4.29 & 0.5332 & 99.7 \\
        ACE $\ell_2$~\cite{jeanneret2023adversarial}  &  6.93 & 22.0 & 100.0& 0.8440    & 1.87  & 2.21  & 0.5946 & 95.0 & 16.4 & 28.2 & 99.6 & 0.7743    & 1.92 & 4.21 & 0.5303 & 95.0 \\ \midrule
        TIME (Ours)                                   & 10.98 & 23.8 & 96.6 & 0.7896    & 2.97  & 2.32  & 0.6303 & 97.1 & 20.9 & 32.9 & 79.3 & 0.6282    & 4.19 & 4.29 & 0.3124 & 89.9 \\ \bottomrule
    \end{tabular}}
    \caption{\textbf{CelebAHQ Evaluation.} \edit{While TIME does not outperform the state-of-the-art metrics, our proposed method provides competitive performance while being completely black-box, \ie having access only to the input and output of the model}. ACE* is \cite{jeanneret2023adversarial}'s method without their post-processing method.}
    \vspace{-3mm}
    \label{tab:celebahq-main}
\end{table*}

\subsection{Quantitative Assessment}

Assessing counterfactuals presents inherent challenges. Despite this, several metrics approximate the core objectives of counterfactual analysis. We will now provide a concise overview of each objective and its frequent evaluation protocol, reserving an in-depth exploration of these metrics for the supplementary material.

\paragraph{Validity.}
First, we need to quantify the ability of the counterfactual explanation method to flip the classifier. 
This is measured by the Success Ratio (SR aka Flip Rate). 

\paragraph{Sparsity and Proximity.}
A counterfactual must have sparse and proximal editions. 
Several metrics have been proposed to evaluate this aspect, depending on the data type. 
For face images\cite{rodriguez2021beyond,jeanneret2023adversarial,Singla2020Explanation,jeanneret2022diffusion}, there are the face verification accuracy (FVA), face similarity (FS), mean number of attributes changed (MNAC), and Correlation Difference (CD). 
For general-purpose images, like BDD100k~\cite{Yu2020BDD100KAD}, the quantitative assessment is done via the SimSiam Similarity ($S^3$)~\cite{jeanneret2023adversarial} and the COUT metric~\cite{khorram2022cycle}. 

\paragraph{Realism.}
The CE research adapts its evaluation metrics from the generation field. 
Hence, the realism of CEs is commonly measured with the FID~\cite{heusel2017gans} and sFID~\cite{jeanneret2023adversarial} metrics but only in the correctly classified images. 

\paragraph{Efficiency.} 
An efficiency analysis is often omitted by many methods. 
A crucial criterion for counterfactual generation techniques is to minimize computation time for generating explanations in ``real time''. 
We evaluate this by contrasting efficiency using floating point operations (FLOPs) per explanation - lower values signify faster inference - and by measuring the average time taken to generate an explanation, specifically within our cluster environment.

\subsubsection{Main Results.} 

Table~\ref{tab:celebahq-main} shows the results of TIME and compares them to the previous literature. 
Although we do not outperform the state-of-the-art in any metric, we found that our results are similar even when our proposed method is restricted to be black-box.
Further, it does not require training of a completely new generative model and does not rely on any optimization for CE generation. 
For the realism metric, we expected to get a low FID~\cite{heusel2017gans} and sFID~\cite{jeanneret2023adversarial} due to the use of Stable Diffusion and beat ACE~\cite{jeanneret2023adversarial}. 
However, ACE uses an inpainting strategy to post-process their counterfactuals.
This reduces this metric because they keep most of the original pixels in their output. 
If we remove the post-processing, the FID increases dramatically. 
With these results, we confirm that T2I generative models are a good tool to explain classifiers counterfactually in a black-box environment.

\subsubsection{Qualitative Results}

We show some qualitative results in Figure~\ref{fig:qualitative} and added more instances in the supplementary material. 
First, we see that DiME~\cite{jeanneret2022diffusion}, ACE~\cite{jeanneret2023adversarial}, and TIME generate very realistic counterfactuals, and the differences are mostly in the details. 
However, the most notable changes are between ACE and our method. 
When we check the regions where ACE made the changes, they are blurred. 
This is due to their over-respacing to create the counterfactual.
For DiME, we checked and found that some of their modifications seem out-of-distribution, for many cases. 
However, TIME produces realistic changes most of the time.
Finally, in our opinion, TIME alterations can be spotted with more ease.

\subsubsection{Efficiency Analysis}

We continue our analysis and study the efficiency of TIME when creating the CE with respect to previous state-of-the-art methods, DiME~\cite{jeanneret2022diffusion} and ACE~\cite{jeanneret2023adversarial}.  
We estimated that TIME uses $98$ TFLOPs and $45$ seconds to create a single counterfactual, using $\tau=35$ as the worst case scenario. 
In contrast, ACE took $279$ TFLOPs and $62$ second per CE while DiME took $1004$ TFLOPs and $163$ seconds.

\subsection{Ablations}

\begin{table}[tb]
    \centering
    \small
    \begin{tabular}{c|c|cccc} \toprule
        Steps               & GS & SR ($\uparrow$)  & FID ($\downarrow$) & FS ($\uparrow$)   & CD ($\downarrow$) \\ \midrule
        \multirow{3}{*}{25} & 3  & 30.1& 35.26& 0.8957& 2.82\\
                            & 4  & 41.0& 30.23& 0.8570& 2.61\\
                            & 5  & 50.1& 27.39& 0.8231& 2.33\\ \midrule
        \multirow{3}{*}{30} & 3  & 62.1& 23.15& 0.8147& 2.34 \\
                            & 4  & 74.0& 22.51& 0.7710& 2.66 \\
                            & 5  & 80.8& 23.51& 0.7300& 2.85 \\ \midrule
        \multirow{3}{*}{35} & 3  & 87.1& 21.69& 0.7227& 2.63 \\
                            & 4  & 92.9& 24.37& 0.6731& 3.03 \\
                            & 5  & 95.0& 27.53& 0.6306& 3.54 \\ \bottomrule
    \end{tabular}
    \caption{\textbf{Steps-Scale trade-off.} We analyze the trade-off between our hyperparameters $\tau$ and $w$. Our results show that increasing $\tau$ gives a strong boost in SR while impacting the other metrics and increasing the generation time. In contrast, $w$ has a similar effect but is less potent without any effect on the generation time.}
    \label{tab:steps-gs}
\end{table}

\begin{table}[tb]
    \centering
    \small
    \begin{tabular}{c|cccc} \toprule
        Context  & SR ($\uparrow$)  & FID ($\downarrow$) & FS ($\uparrow$)   & CD ($\downarrow$)  \\ \midrule
        Without  & 73.9& 23.47& 0.7480 & 2.41 \\
        With     & 92.9& 24.37& 0.6731 & 3.03 \\ \bottomrule
    \end{tabular}
    \caption{\textbf{Context token ablation}. Here, we check the effect of including the context embeddings into our pipeline. The main advantage is increasing the success ratio. This result suggests that we can reduce $\tau$ to reach similar results while being more efficient - less number of EDICT iterations.}
    \label{tab:context}
\end{table}


To show the effectiveness of each component, we realized thorough ablation experiments. 
To this end, we first show the hyperparameter exploration between the depth of the chain of noise $\tau$ and the guidance scale $w$. 
Additionally, we will show the effect of including multiple textual tokens, the context tokens, and, finally, the effect of adding our negative drift – please refer to the practical consideration in section~\ref{sec:gen} for the variable $\tau$. 
Unless explicitly told, we set $\tau=35$ and $w=4$ for all the ablations. 
For the dataset, we did the ablation using 1000 instances of the CelebA HQ validation dataset for the smiling attribute. 
As the quantitative metrics, we used the SR, the FID, the FS, and the CD.

Regarding the FID metric, please note that this metric is very sensible to the number of images. 
When using fewer images, the FID becomes less reliable to compare two methods, and hardly becomes intelligible if the two approaches are evaluated on different number of images. 
Since we use the FID to compare counterfactual on only those instances that flipped the classifier, comparing FIDs where the SR varies significantly does not give any cues.

\paragraph{Steps and scale trade-off.} To begin with, we investigate the effect of the number inversion steps and the scale of the guidance. 
We jointly explore both variables to check the best trade-off, as shown in Table~\ref{tab:steps-gs}.
At first glance, we notice that adding a higher guidance scale or more noise inversion steps produces more successful counterfactuals, assessed with the SR. 
Yet, it comes with a trade-off in other compartments: namely, the quality of the CE, and the amount of editions into the image. 
Generally, increasing $\tau$ or $w$ reflects a decrease in the quality of the image and the increasing numbers of editions. 

\paragraph{Learning the Context Token.} 
Continuing with our study, we analyze the inclusion of our novel context token into our counterfactual generation pipeline. 
To ablate this component, we test whether using our learned context tokens has any advantage in contrast to giving a generic description.
The results are in Table~\ref{tab:context}. 
As we can see, including our tokens provides the best performance gains in terms of SR. 
Qualitatively, the images are similar, yet, the images without context present some artifacts in some cases. 
Furthermore, we see that removing the context provides a boost in the CD and FS metrics. 
Although it seems counterintuitive to include this component, we can easily reach these values by decreasing $\tau$ or $w$ (\eg setting $\tau=30$ and $w=4$, check Table~\ref{tab:steps-gs}), and reducing the inference time.

\paragraph{Effect of the guidance.} We further explore the inclusion of the negative drift term in Eq.~\ref{eq:neg-grads} and show the results in Table~\ref{tab:neg-guidance}. 
From the quantitative assessment, we initially observed that using the classifier-free guidance (CFG in the Table) decreases the SR. 
When denoising the current stage $x_t$ at time $t$, the CFG in Eq.~\ref{eq:class-free} estimates gradients of the log-likelihood conditioned on $C_j$, $-\nabla_{x_t} \log(p(x_t|C_j))$,~\cite{ho2021classifierfree} thus, pushing the generation \textit{toward} the distribution of $C_j$.
In contrast, incorporating the negative guidance (NG) helps steer the generation \textit{away} from the distribution conditioned on $C_i$.
Therefore, the combined effect results in moving the instance from the boundary decision. 
From a qualitative perspective, we did not see major differences.
Nonetheless, as noted in the context of ablation, this can be easily mitigated by reducing $w$ and $\tau$.

\begin{table}[tb]
    \centering
    \small
    \begin{tabular}{c|cccc} \toprule
        Guidance & SR ($\uparrow$)  & FID ($\downarrow$) & FS ($\uparrow$)   & CD ($\downarrow$) \\ \midrule
        CFG      & 75.9& 21.58& 0.7749 & 2.34 \\
        NG       & 92.9& 24.37& 0.6731 & 3.03 \\ \bottomrule
    \end{tabular}
    \caption{\textbf{Negative Guidance}. Here, we check the effect of performing the negative guidance (NG) instead of the classifier-free guidance (CFG). The main advantage is increasing the success ratio. This result suggests that we can reduce $\tau$ to reach similar results while being more efficient.}
    \label{tab:neg-guidance}
\end{table}
\begin{table}[tb]
    \centering
    \small
    \begin{tabular}{c|cccc} \toprule
        Tokens   & SR ($\uparrow$)  & FID ($\downarrow$) & FS ($\uparrow$)   & CD ($\downarrow$)  \\ \midrule
        Single   & 88.1& 22.02& 0.7177 & 3.02 \\
        Multiple & 92.9& 24.37& 0.6731 & 3.03 \\ \bottomrule
    \end{tabular}
    \caption{\textbf{Multiple-tokens Ablation.} We test if using multiple tokens in our pipeline provides any advantage. The results show an increase in SR.}
    \label{tab:multitoken}
\end{table}

\paragraph{Multi-token Inclusion.} Finally, we explore using multiple tokens instead of a single one for both the context and class embeddings, shown in Table~\ref{tab:multitoken}. 
Without any surprise, we noticed that using a single token reduces the SR by a small factor. 
This aligns with the observations given by~\cite{gal2023an}, a token catches enough information of an object or style - or in this case, inductive biases. 
Like in previous analyses, including multiple tokens will increase the efficiency of the model, since we can reach similar performances by tuning $\tau$ or $w$. 
Qualitatively, the most notable change between the images is sharpness.

\paragraph{Recommendations.} Given the previous results, we propose several recommendations for the user and the model debugger, as explained in the introduction. 
Recall that the counterfactual explanations are used as well to recommend changes to the user to get a positive outcome. 
So, for the user, we recommend using the lower amount of iterations $\tau$ and guidance scale $w$. 
This results in a similarity increase and fewer edited characteristics (as evidenced by the CD and FS metrics). 
If the algorithm fails, it is preferable to adjust the guidance scale rather than the number of steps.
For the debugger, always use the context, the negative guidance, and multiple tokens. 
When building the counterfactuals, follow the same recommendations for the user.

\begin{figure}[t]
    \centering
    \includegraphics[width=0.90\columnwidth]{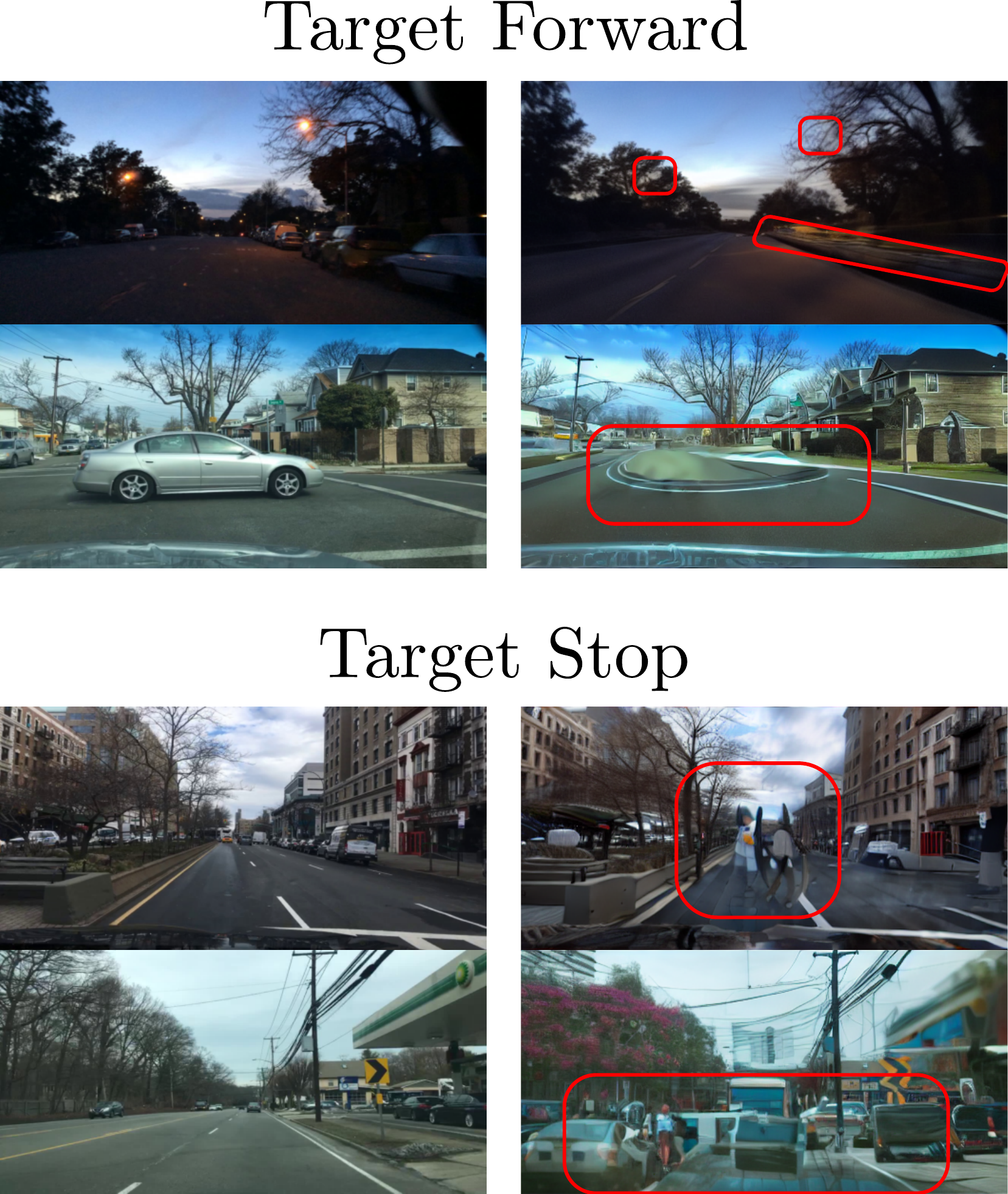}
    \caption{\textbf{BDD100k, a limit for TIME.} TIME changes the entire scene when generating the counterfactuals. Nevertheless, it still gives some insight into what the models have learned, as illustrated by the features inside the red boxes.}
    \label{fig:bdd}
    \vspace{-3mm}
\end{figure}

\begin{table}[t]
    \centering
    \footnotesize
    \resizebox{0.95\columnwidth}{!}{
    \begin{tabular}{c|ccccc} \toprule
        Method        & FID ($\downarrow$)  & sFID ($\downarrow$)  & S$^3$ ($\uparrow$)  & COUT ($\uparrow$)   & SR ($\uparrow$)    \\ \midrule
        STEEX         & 58.8 & -     & -      & -      & 99.5 \\
        DiME          & 7.94 & 11.40 & 0.9463 & 0.2435 & 90.5 \\
        ACE $\ell_1$  & 1.02 & 6.25  & 0.9970 & 0.7451 & 99.9 \\
        ACE $\ell_2$  & 1.56 & 6.53  & 0.9946 & 0.7875 & 99.9 \\\midrule
        TIME (Ours)   & 51.5 & 76.18 & 0.7651 & 0.1490 & 81.8 \\\bottomrule
    \end{tabular}}
    \caption{\textbf{BDD Assessement.} We evaluate the performance of TIME on the complex BDD100k benchmark. On this dataset, there is still room for improvement for black-box counterfactual methods.}
    \label{tab:bdd-main}
\end{table}

\subsection{Limitations.}
To test TIME in more complex scenarios, we generate CEs in the BDD100k~\cite{Yu2020BDD100KAD} dataset using a DenseNet121~\cite{huang2017densely} trained in a \emph{move-forward/stop} binary classification, as in~\cite{jacob2022steex}. 
We show the quantitative evaluation in Table~\ref{tab:bdd-main}.
When generating the explanations, we noticed that TIME modifies most parts of the image, unfortunately, as shown by the $S^3$ metric. 
This is expected, as this task is challenging since it requires multiple factors to decide if to stop or to move forward. 
Nevertheless, we believe that these explanations still give some useful insights as a debugging tool. 
For example, Figure~\ref{fig:bdd} shows that removing the red lights and adding motion blur will change the classification from \emph{stop} to \emph{move}, as evidenced in~\cite{jeanneret2023adversarial}, or adding objects in front will flip the prediction to \emph{stop}. 

We believe that counterfactual methods for tasks dependent on complex scenes, where the decision is impacted by large objects or co-occurrences of several stimuli, require specific architectures. 
In fact, we noticed that ACE~\cite{jeanneret2023adversarial} mainly adds some small modifications (\eg changing the red lights), which is not inaccurate but is too constrained and cannot explore more insights about the learned features. 
Indeed, the work of Zemni~\etal~\cite{zemni2023octet} focuses only on the object aspect of counterfactuals, in this case using an object-centric generator, BlobGAN~\cite{epstein2022blobgan}. 
This suggests that general-purpose counterfactual methods are not adapted for these tasks.

\section{Conclusion}

In this work, we present TIME, a counterfactual generation method to analyze classifiers disregarding their architecture and weights, only by looking at their inputs and outputs. 
By leveraging T2I generative models and a distillation approach, our method is capable of producing CEs for black-box models, a complex scenario not tackled before. 
Further, we show the advantages and limitations of TIME and shed light on possible future works.
We believe that our approach opens the door to research focus on counterfactual methods in the challenging scenario of the black-box models.

\textbf{Acknowledgements} 
Research reported in this publication was supported by the Agence Nationale pour la Recherche (ANR) under award number ANR-19-CHIA-0017.

{\small
\bibliographystyle{ieee_fullname}
\bibliography{egbib}
}

\clearpage
\appendix

\onecolumn


\parbox[t][3cm][c]{0.95\textwidth}{\centering \Large \textbf{Supplementary Material\\Text-to-Image Models for Counterfactual Explanations: a Black-Box Approach}}

\section{Evaluation Criteria}

Before describing each metric and its formulation, we will thoroughly describe the goals of counterfactual explanations. 
As we stated in the main manuscript, counterfactual explanations seek to change an instance prediction by modifying the input instance. 
However, these modifications must be small but perceptually coherent. 
From the previous statement, we can extract many goals of CEs: 
\begin{enumerate}
    \item CEs must flip the decision of the classifier. In the literature, this feature is called \textit{validity}.
    \item The counterfactual changes should be plausible and realistic - simply referred to as \textit{realism}.
    Visual automated systems are generally brittle to adversarial noise~\cite{goodfellow2014explaining}. 
    This noise is designed to fool the classifier, but with the restriction that it is hidden from visual inspection. 
    Since this noise cannot be perceived, it cannot be analyzed to find spurious correlations. 
    Therefore, only realistic and plausible changes are allowed. 
    \item The algorithm must generate \textit{proximal and sparse} counterfactuals. 
    One could create a valid and realistic explanation by simply replacing the target instance with a new one. 
    This still obeys the realistic and valid goals. 
    However, it does not give any information about the variables. 
    Thus, the modifications must be sparse and close to the image to visually observe which variables have changed.
    \item Finally, the algorithm must generate the explanation \textit{efficiently}. 
    This property is required to avoid delays for the user.
\end{enumerate}

Now we will proceed to describe each evaluation metric and link it to its corresponding objective. 
As for notations, let $M(x, y)$ be the counterfactual algorithm applied to an image $x \in D$ targeting the class $y$, where $D$ is a dataset. 
Additionally, let $C$ be the classifier, $\mathbb{1}(condition)$ a function that is one if the condition is true or zero otherwise. 
Finally, let $a\in A$ be an attribute in a set $A$, then $O^a$ is an attribute oracle classifier for $a$.
This network predicts if its input has the attribute $a$.
Similarly, let $\mathcal{O}$ be an identity verification network
This DNN is trained to give a similarity measure between two images, often computed with the cosine similarity $CS$.

\paragraph{Success Rate.} The success rate (or flip rate) measures the ratio at which counterfactuals have successfully reversed the original classifier's decision. 
This metric correlates with the validity goal. 
To measure it, we simply compute the proportion of valid counterfactuals to the size of the dataset, as in
\begin{equation}
    SR = \frac{1}{|D|} \sum_{x \in D} \mathbb{1}(C(M(x, y)) = y).
\end{equation}

\paragraph{Realism.} 
To approximate the realism of the counterfactuals, the literature adopts the FID~\cite{heusel2017gans} metric from generation research. 
Furthermore, \cite{jeanneret2023adversarial} extended the metric by computing the FID between the half of the dataset and the counterfactuals of the complement set. 
This was motivated to reduce the inherent bias in computing the FID, given that the difference between the original images and their CE is a few pixels in the image.

\paragraph{Proximity and Sparsity.} 
To evaluate this goal, previous methods proposed several metrics to quantify the degree of dissimilarity between an instance and its explanation. 
Initially, most metrics were proposed for face images. 
Initially, \cite{jacob2022steex} suggested using the mean number of attributes changed (MNAC), computed as follows:
\begin{equation}
    MNAC = \frac{1}{|D|} \sum_{x \in D} \sum_{a \in A} \mathbb{1}(O^a(M(x, y)) \neq O^a(x)).
\end{equation}
However, \cite{jeanneret2022diffusion} noted that counterfactual methods will change some attributes if they are correlated. 
Thus, based on the MNAC, the Correlation Difference (CD)~\cite{jeanneret2022diffusion} measures the correlations produced by $M$.
To further assess the proximity and sparsity in face counterfactuals, \cite{Singla2020Explanation} suggested using the Face Verification Accuracy (FVA) to compute whether $M$ cannot modify the identity of the person. 
This metric is calculated as
\begin{equation}
    FVA = \frac{1}{|D|} \sum_{x \in D} \mathbb{1} (CS(\mathcal{O}(x), \mathcal{O}(M(x,y))) > 0.5).
\end{equation}
\cite{jeanneret2023adversarial} noted that this metric was already saturated.
To measure a more fine-grained metric, they proposed taking the continuous $CS$ and calling the metric face similarity (FS):
\begin{equation}\label{eq:fs}
    FS = \frac{1}{|D|} \sum_{x \in D} CS(\mathcal{O}(x), \mathcal{O}(M(x,y)).
\end{equation}
Finally, the same authors extended this metric for general-purpose images by computing Eq.~\ref{eq:fs} using a self-supervised trained model as $\mathcal{O}$. 
They called this metric $S^3$.
Finally, \cite{khorram2022cycle} proposed to compute COUT. 
This metric computes the probability of the class $y$ using multiple linear interpolations between $x$ and $M(x, y)$.

\paragraph{Efficiency.}
The literature generally ignores computing an \textit{efficiency} metric. 
To compute the efficiency of counterfactual models, the widely accepted metric is floating point operations (FLOPs). 
In addition, it is also recommended to compute the average time per counterfactual. 
However, this metric is only comparable if all measurements are computed on the under the same circumstances. 
\section{Qualitative Results}

In this section, we provide additional qualitative results. 
For the CelebA HQ~\cite{CelebAMask-HQ} dataset, we provide our and ACE~\cite{jeanneret2023adversarial} counterfactuals to show the differences.

\begin{figure*}[b]
    \centering
    \includegraphics[width=0.95\textwidth]{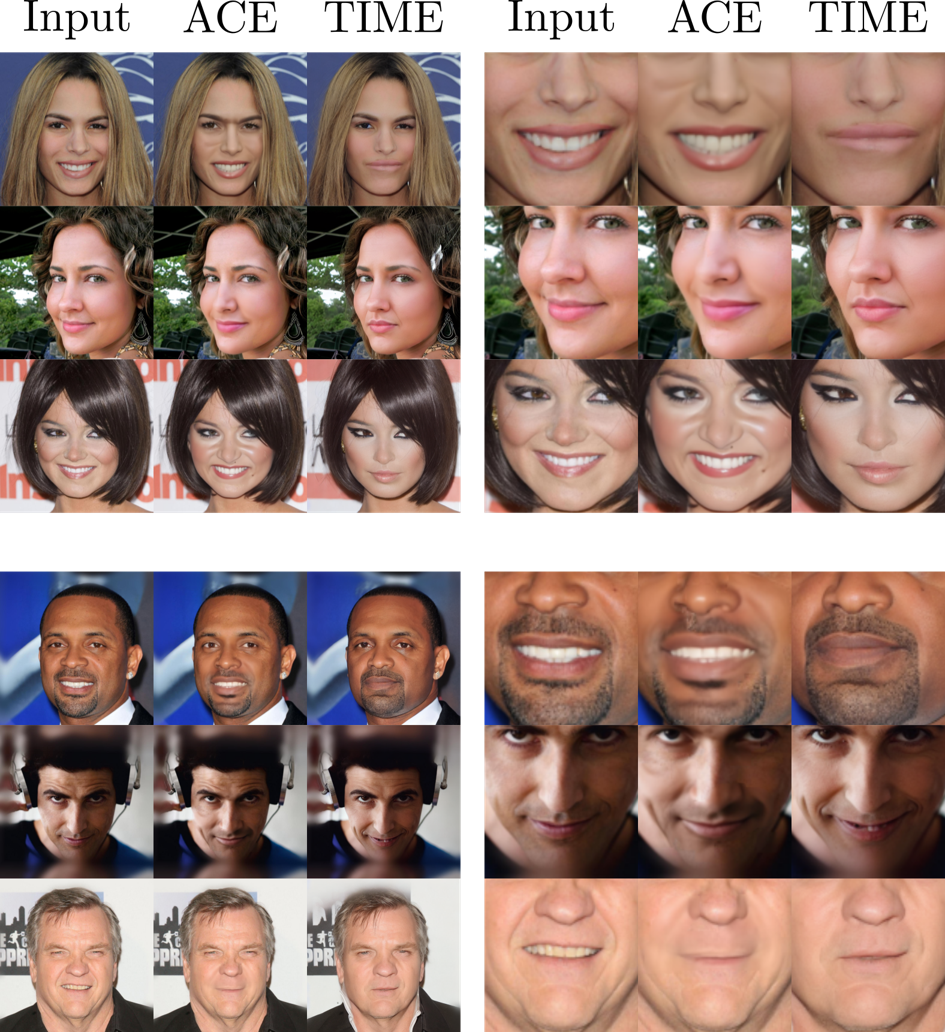}
    \caption{Counterfactual Explanations targeting the Non-Smile attribute.}
    \label{fig:nonsmile}
\end{figure*}

\begin{figure*}
    \centering
    \includegraphics[width=0.95\textwidth]{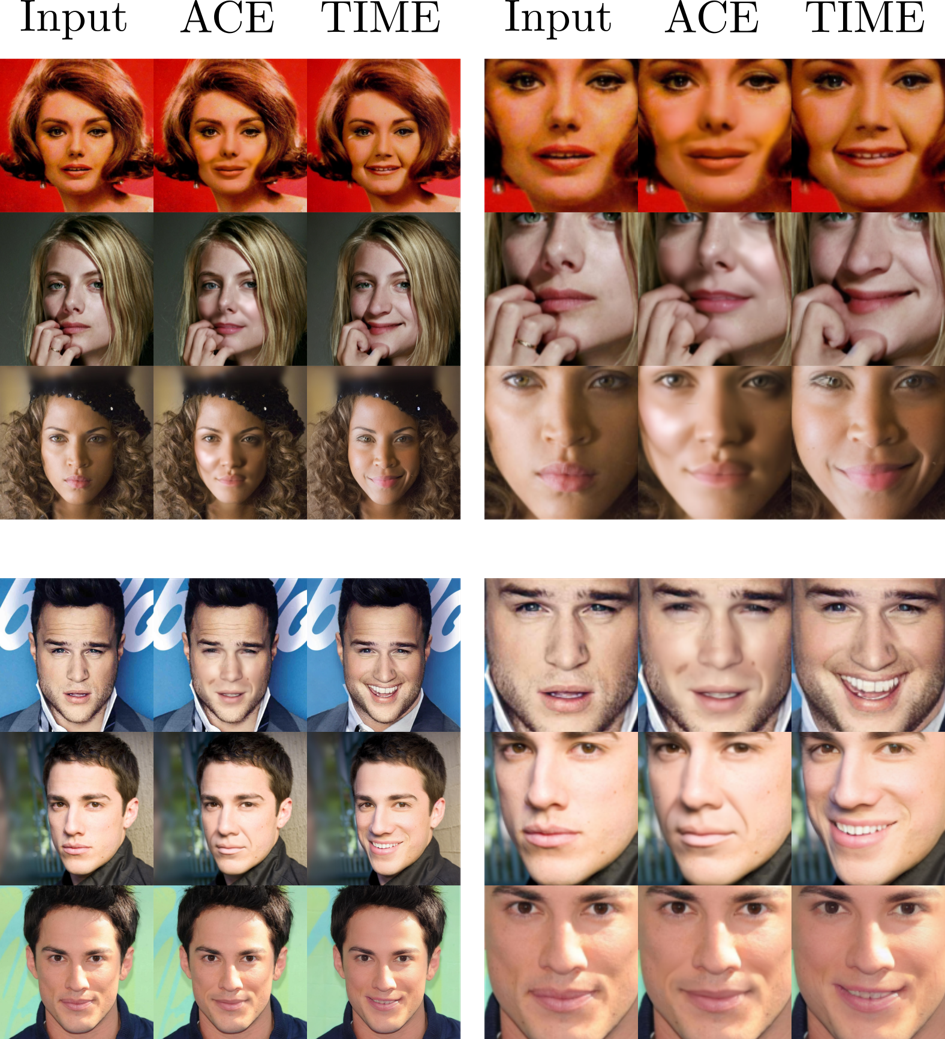}
    \caption{Counterfactual Explanations targeting the Smile attribute.}
    \label{fig:smile}
\end{figure*}

\begin{figure*}
    \centering
    \includegraphics[width=0.95\textwidth]{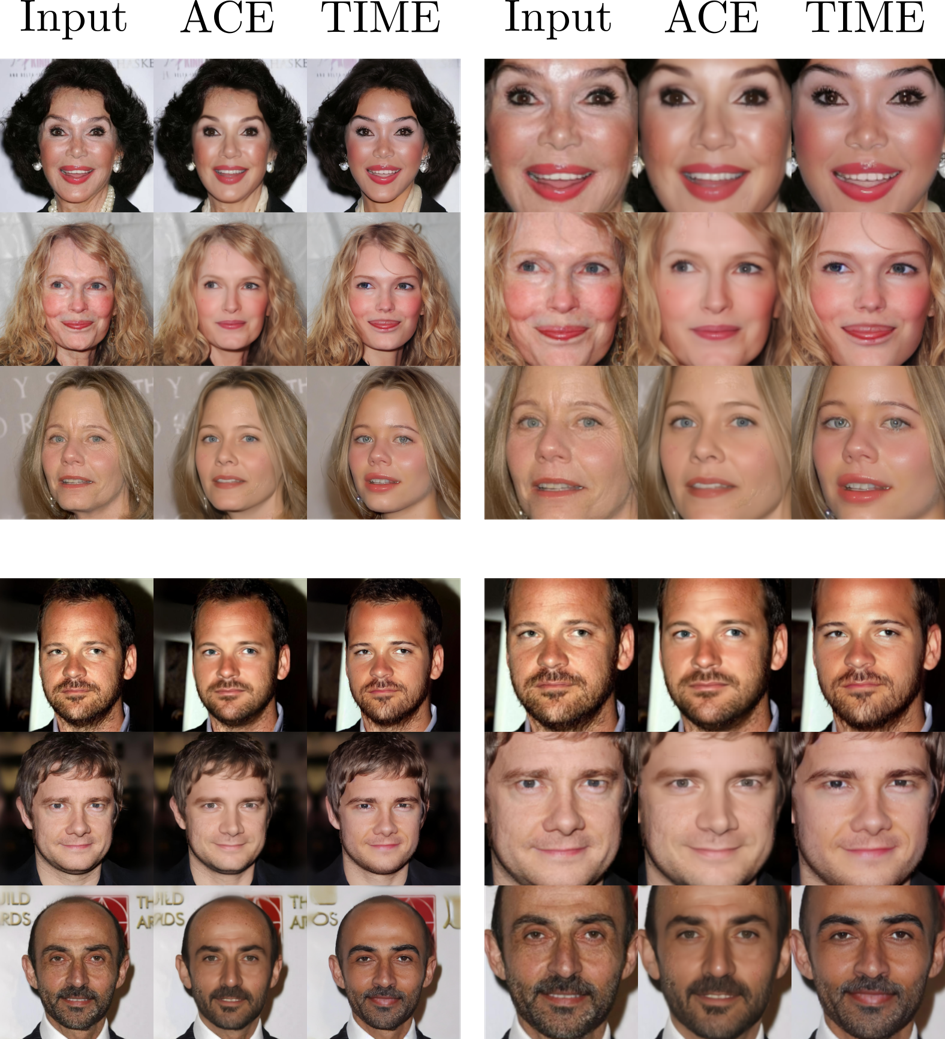}
    \caption{Counterfactual Explanations targeting the Young attribute.}
    \label{fig:young}
\end{figure*}

\begin{figure*}
    \centering
    \includegraphics[width=0.95\textwidth]{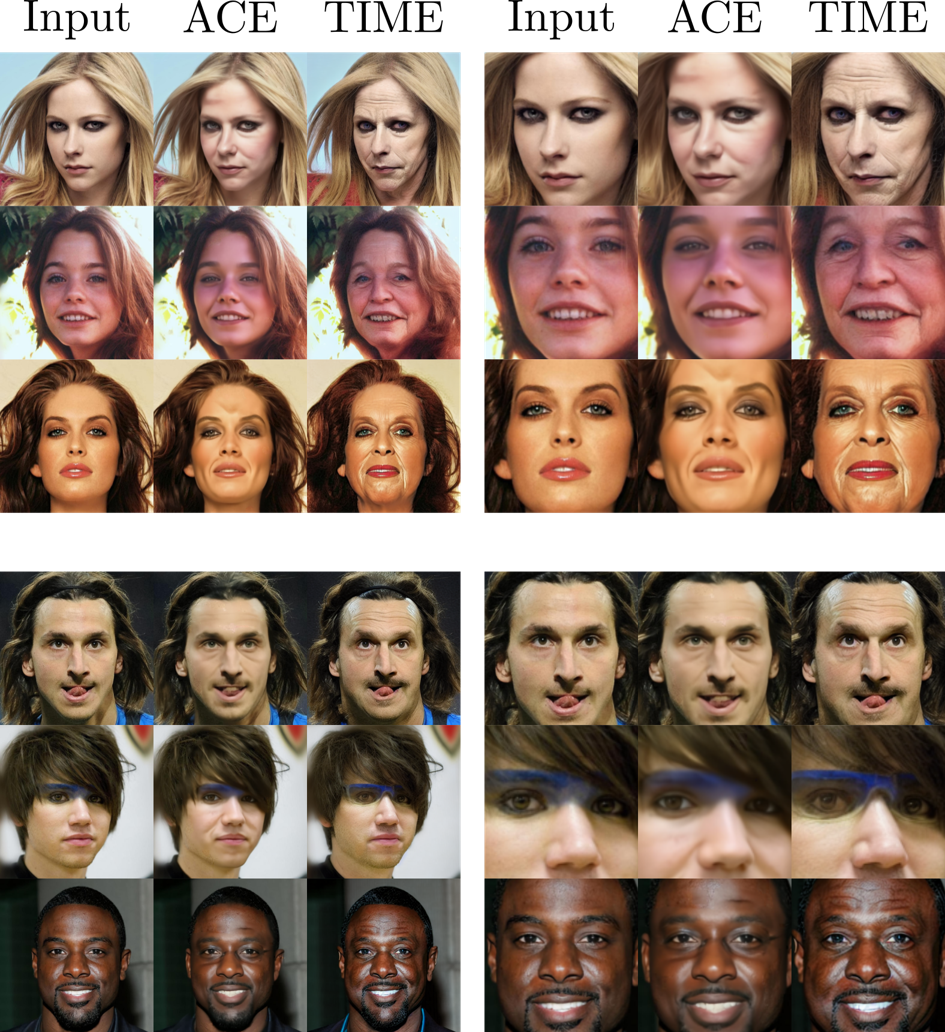}
    \caption{Counterfactual Explanations targeting the Old attribute.}
    \label{fig:old}
\end{figure*}

\begin{figure*}
    \centering
    \includegraphics[width=0.95\textwidth]{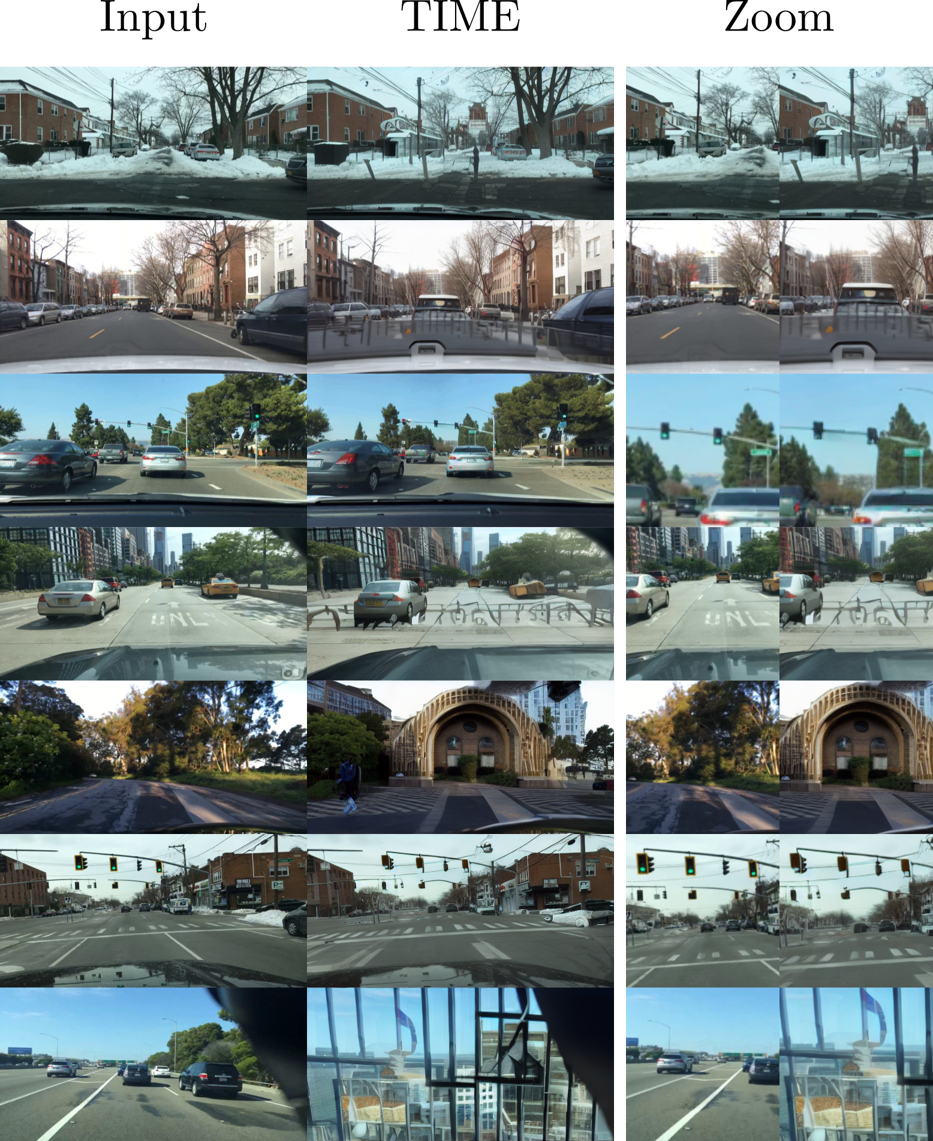}
    \caption{Counterfactual Explanations targeting the Stop action.}
    \label{fig:stop}
\end{figure*}

\begin{figure*}
    \centering
    \includegraphics[width=0.95\textwidth]{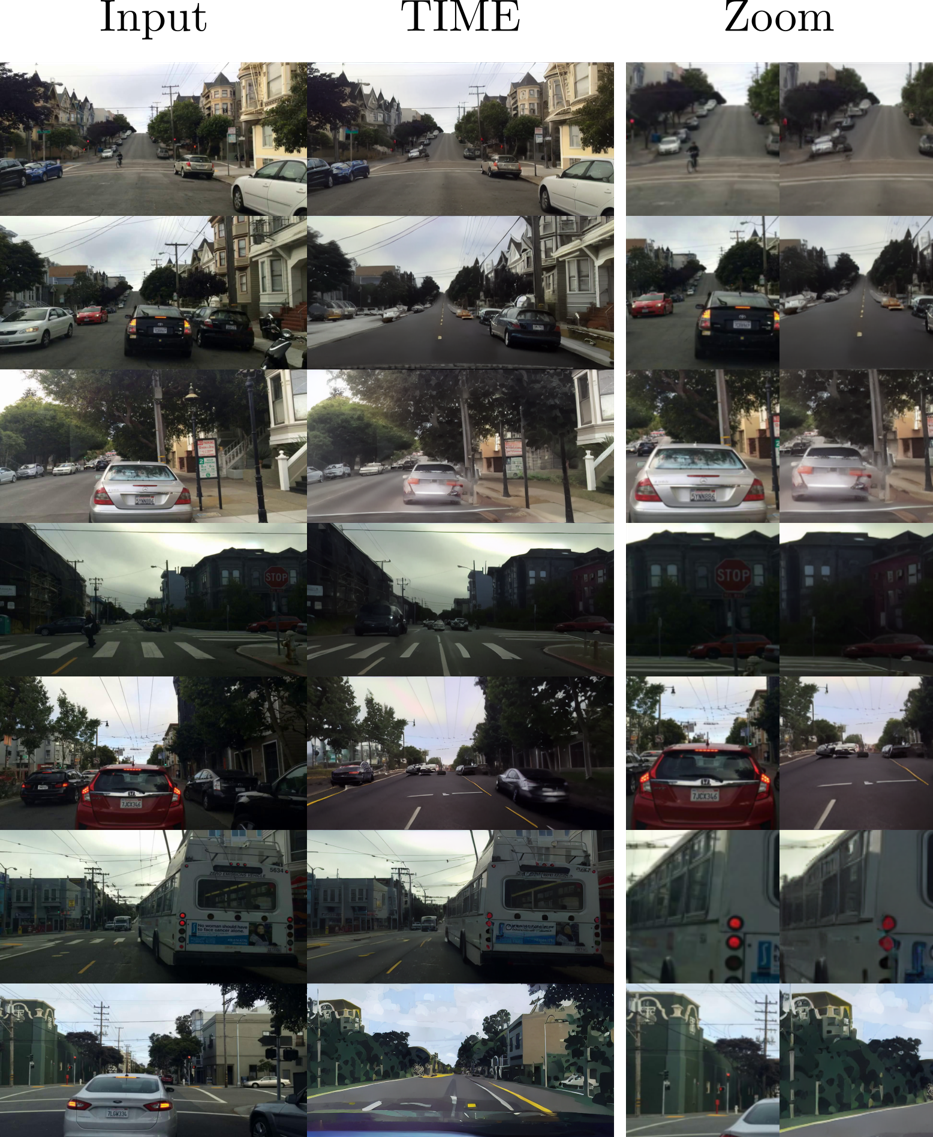}
    \caption{Counterfactual Explanations targeting the Forward action.}
    \label{fig:stop}
\end{figure*}

\end{document}